\def\year{2019}\relax
\documentclass[letterpaper]{article} 
\usepackage{aaai19}  
\usepackage{times}  
\usepackage{helvet}  
\usepackage{courier}  
\usepackage{url}  
\usepackage{graphicx}  

\usepackage{algorithm}
\usepackage{algpseudocode}
\usepackage{amsfonts}
\usepackage{amsmath}
\usepackage{bm}
\usepackage{booktabs}
\usepackage{braket}
\usepackage{mathtools}
\usepackage[subrefformat=parens]{subcaption}

\DeclareMathOperator*{\argmax}{arg\,max}

\begin{document}
%
\title{Variational Autoencoder with Implicit Optimal Priors}
\author{
Hiroshi Takahashi$^1$,
Tomoharu Iwata$^2$,
Yuki Yamanaka$^3$,
Masanori Yamada$^3$,
Satoshi Yagi$^1$
\\
$^1$ NTT Software Innovation Center\\
$^2$ NTT Communication Science Laboratories\\
$^3$ NTT Secure Platform Laboratories\\
\{takahashi.hiroshi,
iwata.tomoharu,
yamanaka.yuki,
yamada.m,
yagi.satoshi\}@lab.ntt.co.jp
}
\nocopyright
\maketitle

\begin{abstract}
  The variational autoencoder (VAE) is a powerful generative model that can estimate the probability of a data point by using latent variables.
  In the VAE, the posterior of the latent variable given the data point is regularized by the prior of the latent variable using Kullback Leibler (KL) divergence.
  Although the standard Gaussian distribution is usually used for the prior,
  this simple prior incurs over-regularization.
  As a sophisticated prior, the aggregated posterior has been introduced,
  which is the expectation of the posterior over the data distribution.
  This prior is optimal for the VAE in terms of maximizing the training objective function.
  However, KL divergence with the aggregated posterior cannot be calculated in a closed form,
  which prevents us from using this optimal prior.
  With the proposed method, we introduce the density ratio trick to estimate this KL divergence without modeling the aggregated posterior explicitly.
  Since the density ratio trick does not work well in high dimensions,
  we rewrite this KL divergence that contains the high-dimensional density ratio
  into the sum of the analytically calculable term and the low-dimensional density ratio term,
  to which the density ratio trick is applied.
  Experiments on various datasets show that the VAE with this implicit optimal prior achieves high density estimation performance.
\end{abstract}

\section{Introduction}
Estimating data distributions is one of the important challenges of machine learning.
The variational autoencoder (VAE) \cite{kingma2013auto,rezende2014stochastic} was presented as a powerful generative model
that can learn distributions by using latent variables and neural networks.
Since the VAE can capture the high-dimensional complicated data distributions,
it is widely applied to various data, such as images \cite{gulrajani2016pixelvae}, videos \cite{gregor2015draw}, and audio and speech \cite{hsu2017learning,van2017neural}.

The VAE is composed of three distributions: the encoder, the decoder, and the prior of the latent variable.
The encoder and the decoder are conditional distributions,
and neural networks are used to model these distributions.
The encoder defines the posterior of the latent variable given the data point,
whereas the decoder defines the distribution of the data point given the latent variable.
The parameters of encoder and decoder neural networks are optimized by maximizing the sum of the evidence lower bound of the log marginal likelihood.
In the training of VAE,
the prior regularizes the encoder by Kullback Leibler (KL) divergence.
The standard Gaussian distribution is usually used for the prior since the KL divergence can be calculated in a closed form.

Recent research shows that the prior plays an important role in the density estimation \cite{hoffman2016elbo}.
Although the standard Gaussian prior is usually used,
this simple prior incurs over-regularization,
which is one of the causes of the poor density estimation performance.
This over-regularization is also known as the posterior-collapse phenomenon \cite{van2017neural}.
To improve the density estimation performance,
the aggregated posterior prior has been introduced, which is the expectation of the encoder over the data distribution \cite{hoffman2016elbo}.
The aggregated posterior is an optimal prior in terms of maximizing the training objective function of the VAE.
However, KL divergence with the aggregated posterior cannot be calculated in a closed form,
which prevents us from using this optimal prior.
In previous work \cite{tomczak2017vae}, the aggregated posterior is modeled by using the finite mixture of encoders
for calculating the KL divergence in a closed form.
Nevertheless, it has sensitive hyperparameters such as the number of mixture components,
which are difficult to tune.

In this paper, we propose the VAE with implicit optimal priors,
where the aggregated posterior is used as the prior,
but the KL divergence is directly estimated without modeling the aggregated posterior explicitly.
This implicit modeling enables us to avoid the difficult hyperparameter tuning for the aggregated posterior model.
We use the density ratio trick, which can estimate the density ratio between two distributions without modeling each distribution explicitly,
since the KL divergence is the expectation of the density ratio between the encoder and aggregated posterior.
Although the density ratio trick is powerful,
it has been experimentally shown to work poorly in high dimensions \cite{sugiyama2012density,rosca2018distribution}.
Unfortunately, with high-dimensional datasets, the density ratio between the encoder and the aggregated posterior also becomes high-dimensional.
To avoid the density ratio estimation in high dimensions,
we rewrite the KL divergence with the aggregated posterior to the sum of two terms.
The first term is the KL divergence between the encoder and the standard Gaussian prior, which can be calculated in a closed form.
The other term is the low-dimensional density ratio between the aggregated posterior and the standard Gaussian distribution,
to which the density ratio trick is applied.

\section{Preliminaries}

\subsection{Variational Autoencoder}
\label{sec:vae}

First, we review the variational autoencoder (VAE) \cite{kingma2013auto,rezende2014stochastic}.
The VAE is a probabilistic latent variable model that relates an observed variable vector $\mathbf{x}$ to a low-dimensional latent variable vector
$\mathbf{z}$ by a conditional distribution.
The VAE models the probability of a data point $\mathbf{x}$ by
\begin{align}
  p_{\mathbf{\theta}}(\mathbf{x})&=\int p_{\mathbf{\theta}}(\mathbf{x}\mid\mathbf{z})p_{\mathbf{\lambda}}(\mathbf{z})\mathrm{d}\mathbf{z},
\end{align}
where $p_{\mathbf{\lambda}}(\mathbf{z})$ is a prior of the latent variable vector,
and $p_{\mathbf{\theta}}(\mathbf{x}|\mathbf{z})$ is the conditional distribution of $\mathbf{x}$ given $\mathbf{z}$,
which is modeled by neural networks with parameter $\theta$.
For example, if $\mathbf{x}$ is binary, this distribution is modeled by a Bernoulli distribution $\mathcal{B}(\mathbf{x}\mid\mu_{\theta}(\mathbf{z}))$,
where $\mu_{\theta}(\mathbf{z})$ is neural networks with parameter $\theta$ and input $\mathbf{z}$.
These neural networks are called the decoder.

The log marginal likelihood $\ln p_{\mathbf{\theta}}(\mathbf{x})$ is bounded below by the evidence lower bound (ELBO),
which is derived from Jensen's inequality, as follows:
\begin{align}
  \ln p_{\mathbf{\theta}}(\mathbf{x})&=\ln\mathbb{E}_{q_{\mathbf{\phi}}(\mathbf{z}\mid\mathbf{x})}\left[\frac{p_{\mathbf{\theta}}(\mathbf{x}\mid\mathbf{z})p_{\mathbf{\lambda}}(\mathbf{z})}{q_{\mathbf{\phi}}(\mathbf{z}\mid\mathbf{x})}\right]\nonumber\\
  &\geq\mathbb{E}_{q_{\mathbf{\phi}}(\mathbf{z}\mid\mathbf{x})}\left[\ln\frac{p_{\mathbf{\theta}}(\mathbf{x}\mid\mathbf{z})p_{\mathbf{\lambda}}(\mathbf{z})}{q_{\mathbf{\phi}}(\mathbf{z}\mid\mathbf{x})}\right]\nonumber\\
  &\equiv\mathcal{L}(\mathbf{x};\mathbf{\theta},\mathbf{\phi}),
  \label{eq:jensen}
\end{align}
where $\mathbb{E}[\cdot]$ represents the expectation, and $q_{\mathbf{\phi}}(\mathbf{z}\mid\mathbf{x})$ is the posterior of $\mathbf{z}$ given $\mathbf{x}$,
which are modeled by neural networks with parameter $\phi$.
$q_{\mathbf{\phi}}(\mathbf{z}\mid\mathbf{x})$ is usually modeled by a Gaussian distribution $\mathcal{N}(\mathbf{z}\mid\mu_{\phi}(\mathbf{x}),\sigma_{\phi}^{2}(\mathbf{x}))$,
where $\mu_{\phi}(\mathbf{x})$ and $\sigma_{\phi}^{2}(\mathbf{x})$ are neural networks with parameter $\phi$ and input $\mathbf{x}$.
These neural networks are called the encoder.

The ELBO (Eq. (\ref{eq:jensen})) can be also written as
\begin{multline}
  \mathcal{L}(\mathbf{x};\mathbf{\theta},\mathbf{\phi})=-D_{KL}(q_{\mathbf{\phi}}(\mathbf{z}\mid\mathbf{x})\Vert p_{\mathbf{\lambda}}(\mathbf{z}))\\
  +\mathbb{E}_{q_{\mathbf{\phi}}(\mathbf{z}\mid\mathbf{x})}\left[\ln p_{\mathbf{\theta}}(\mathbf{x}\mid\mathbf{z})\right],
  \label{eq:elbo}
\end{multline}
where $D_{KL}(P \Vert Q)$ is the Kullback Leibler (KL) divergence between $P$ and $Q$.
The second expectation term in Eq. (\ref{eq:elbo}) is called the reconstruction term,
which is also known as the negative reconstruction error.

The parameters of the encoder and decoder neural networks are optimized
by maximizing the following expectation of the lower bound of the log marginal likelihood:
\begin{align}
  \max_{\mathbf{\theta},\mathbf{\phi}}\int p_{\mathcal{D}}(\mathbf{x})\mathcal{L}(\mathbf{x};\mathbf{\theta},\mathbf{\phi})\mathrm{d}\mathbf{x},
  \label{eq:objective}
\end{align}
where $p_{\mathcal{D}}(\mathbf{x})$ is the data distribution.

\subsection{Aggregated Posterior}
\label{sec:ap}

The training of VAE is maximizing the reconstruction term with regularization by KL divergence between the encoder and the prior.
The prior is usually modeled by a standard Gaussian distribution $\mathcal{N}(\mathbf{z}|\mathbf{0},\mathbf{I})$ \cite{kingma2013auto}.
However, this is not an optimal prior for the VAE.
This simple prior incurs over-regularization,
which is one of the causes of the poor density estimation performance \cite{hoffman2016elbo}.
This phenomenon is called the posterior-collapse \cite{van2017neural}.

The optimal prior that maximizes the objective function of VAE (Eq. (\ref{eq:objective})) can be derived analytically.
The maximization of Eq. (\ref{eq:objective}) with respect to the prior $p_{\mathbf{\lambda}}(\mathbf{z})$ is written as follows:
\begin{align}
  &\argmax_{p_{\mathbf{\lambda}}(\mathbf{z})}\int p_{\mathcal{D}}(\mathbf{x})\mathcal{L}(\mathbf{x};\mathbf{\theta},\mathbf{\phi})\mathrm{d}\mathbf{x}\nonumber\\
  &=\argmax_{p_{\mathbf{\lambda}}(\mathbf{z})}\int p_{\mathcal{D}}(\mathbf{x})\mathbb{E}_{q_{\mathbf{\phi}}(\mathbf{z}\mid\mathbf{x})}\left[\ln p_{\mathbf{\lambda}}(\mathbf{z})\right]\mathrm{d}\mathbf{x}\nonumber\\
  &=\argmax_{p_{\mathbf{\lambda}}(\mathbf{z})}\int\left\{ \int p_{\mathcal{D}}(\mathbf{x})q_{\mathbf{\phi}}(\mathbf{z}\mid\mathbf{x})\mathrm{d}\mathbf{x}\right\} \ln p_{\mathbf{\lambda}}(\mathbf{z})\mathrm{d}\mathbf{z}\nonumber\\
  &=\argmax_{p_{\mathbf{\lambda}}(\mathbf{z})}-H(\int p_{\mathcal{D}}(\mathbf{x})q_{\mathbf{\phi}}(\mathbf{z}\mid\mathbf{x})\mathrm{d}\mathbf{x},p_{\mathbf{\lambda}}(\mathbf{z})),
  \label{eq:cross_entropy}
\end{align}
where $-H(P, Q)$ is the negative cross entropy between $P$ and $Q$.
Since $-H(P, Q)$ takes a maximum value when $P$ is equal to $Q$,
the optimal prior $p_{\mathbf{\lambda}}^{\ast}(\mathbf{z})$ that maximizes Eq. (\ref{eq:objective}) is
\begin{align}
  p_{\mathbf{\lambda}}^{\ast}(\mathbf{z})=\int p_{\mathcal{D}}(\mathbf{x})q_{\mathbf{\phi}}(\mathbf{z}\mid\mathbf{x})\mathrm{d}\mathbf{x}\equiv q_{\mathbf{\phi}}(\mathbf{z}).
\end{align}
This distribution $q_{\mathbf{\phi}}(\mathbf{z})$ is called the aggregated posterior.

When we use the standard Gaussian prior $p(\mathbf{z})=\mathcal{N}(\mathbf{z}|\mathbf{0},\mathbf{I})$,
the KL divergence $D_{KL}(q_{\mathbf{\phi}}(\mathbf{z}\mid\mathbf{x})\Vert p(\mathbf{z}))$ can be calculated in a closed form \cite{kingma2013auto}.
However, when we use the aggregated posterior $q_{\mathbf{\phi}}(\mathbf{z})$ as the prior,
the KL divergence
\begin{align}
  D_{KL}(q_{\mathbf{\phi}}(\mathbf{z}\mid\mathbf{x})\Vert q_{\mathbf{\phi}}(\mathbf{z}))&=\mathbb{E}_{q_{\mathbf{\phi}}(\mathbf{z}\mid\mathbf{x})}\left[\ln\frac{q_{\mathbf{\phi}}(\mathbf{z}\mid\mathbf{x})}{q_{\mathbf{\phi}}(\mathbf{z})}\right]
  \label{eq:kl_divergence}
\end{align}
cannot be calculated in a closed form,
which prevents us from using the aggregated posterior as the prior.

\subsection{Previous work: VampPrior}
\label{sec:vamp}

In previous work, the aggregated posterior is modeled by using the finite mixture of encoders to calculate the KL divergence.
Given a dataset $\mathbf{X}=\left\{\mathbf{x}^{(1)},\ldots,\mathbf{x}^{(N)}\right\}$,
the aggregated posterior can be simply modeled by an empirical distribution:
\begin{align}
  q_{\mathbf{\phi}}(\mathbf{z})&\simeq\frac{1}{N}\sum_{i=1}^{N}q_{\mathbf{\phi}}(\mathbf{z}\mid\mathbf{x}^{(i)}).
  \label{eq:empirical}
\end{align}
Nevertheless, this empirical distribution incurs over-fitting \cite{tomczak2017vae}.
Thus, the VampPrior \cite{tomczak2017vae} models the aggregated posterior by
\begin{align}
  q_{\mathbf{\phi}}(\mathbf{z})&\simeq\frac{1}{K}\sum_{k=1}^{K}q_{\mathbf{\phi}}(\mathbf{z}\mid\mathbf{u}^{(k)}),
  \label{eq:vamp}
\end{align}
where $K$ is the number of mixtures,
and $\mathbf{u}^{(k)}$ is the same dimensional vector as a data point.
$\mathbf{u}$ is regarded as the pseudo input for the encoder,
and is optimized during the training of the VAE through the stochastic gradient descent (SGD).
If $K \ll N$, the VampPrior can avoid over-fitting \cite{tomczak2017vae}.
The KL divergence with the VampPrior can be calculated by the Monte Carlo approximation.
The VAE with the VampPrior achieves better density estimation performance than the VAE with the standard Gaussian prior and the VAE with the Gaussian mixture prior \cite{dilokthanakul2016deep}.
However, this approach has a major drawback:
it has sensitive hyperparameters such as the number of mixtures $K$,
which are difficult to tune.

From the above discussion,
the aggregated posterior seems to be difficult to model explicitly.
In this paper, we estimate the KL divergence with the aggregated posterior without modeling the aggregated posterior explicitly.

\section{Proposed Method}

In this section, we propose the approximation method of the KL divergence with the aggregated posterior,
and describe the optimization procedure of our approach.

\subsection{Estimating the KL Divergence}
\label{sec:kl}

As shown in Eq. (\ref{eq:kl_divergence}),
the KL divergence with the aggregated posterior is the expectation of the logarithm of the density ratio $q_{\mathbf{\phi}}(\mathbf{z}\mid\mathbf{x})/q_{\mathbf{\phi}}(\mathbf{z})$.
In this paper, we introduce the density ratio trick \cite{sugiyama2012density,goodfellow2014generative},
which can estimate the ratio of two distributions without modeling each distribution explicitly.
Hence, there is no need to model the aggregated posterior explicitly.
By using the density ratio trick,
$q_{\mathbf{\phi}}(\mathbf{z}\mid\mathbf{x})/q_{\mathbf{\phi}}(\mathbf{z})$
can be estimated by using a probabilistic binary classifier $D(\mathbf{x},\mathbf{z})$.

However, the density ratio trick has a serious drawback:
it has been experimentally shown to work poorly in high dimensions \cite{sugiyama2012density,rosca2018distribution}.
Unfortunately, if $\mathbf{x}$ is high-dimensional,
$q_{\mathbf{\phi}}(\mathbf{z}\mid\mathbf{x})/q_{\mathbf{\phi}}(\mathbf{z})$ also becomes a high-dimensional density ratio.
The reason is as follows.
Since the $q_{\mathbf{\phi}}(\mathbf{z}\mid\mathbf{x})$ is a conditional distribution of $\mathbf{z}$ given $\mathbf{x}$,
the density ratio trick has to use a probabilistic binary classifier $D(\mathbf{x},\mathbf{z})$,
which takes $\mathbf{x}$ and $\mathbf{z}$ jointly as an input.
In fact, $D(\mathbf{x},\mathbf{z})$ estimates the density ratio of joint distributions of $\mathbf{x}$ and $\mathbf{z}$,
which is a high-dimensional density ratio with high-dimensional $\mathbf{x}$ \cite{mescheder2017adversarial}.

To avoid the density ratio estimation in high dimensions,
we rewrite the KL divergence $D_{KL}(q_{\mathbf{\phi}}(\mathbf{z}\mid\mathbf{x})\Vert q_{\mathbf{\phi}}(\mathbf{z}))$ as follows:
\begin{align}
  &D_{KL}(q_{\mathbf{\phi}}(\mathbf{z}\mid\mathbf{x})\Vert q_{\mathbf{\phi}}(\mathbf{z}))\nonumber\\
  &=\mathbb{E}_{q_{\mathbf{\phi}}(\mathbf{z}\mid\mathbf{x})}\left[\ln\frac{q_{\mathbf{\phi}}(\mathbf{z}\mid\mathbf{x})}{q_{\mathbf{\phi}}(\mathbf{z})}\right]\nonumber\\
  &=\int q_{\mathbf{\phi}}(\mathbf{z}\mid\mathbf{x})\ln\frac{q_{\mathbf{\phi}}(\mathbf{z}\mid\mathbf{x})}{p(\mathbf{z})}\mathrm{d}\mathbf{z}\nonumber\\
  &\qquad\qquad\qquad\qquad\quad+\int q_{\mathbf{\phi}}(\mathbf{z}\mid\mathbf{x})\ln\frac{p(\mathbf{z})}{q_{\mathbf{\phi}}(\mathbf{z})}\mathrm{d}\mathbf{z}\nonumber\\
  &=D_{KL}(q_{\mathbf{\phi}}(\mathbf{z}\mid\mathbf{x})\Vert p(\mathbf{z}))-\mathbb{E}_{q_{\mathbf{\phi}}(\mathbf{z}\mid\mathbf{x})}\left[\ln\frac{q_{\mathbf{\phi}}(\mathbf{z})}{p(\mathbf{z})}\right].
  \label{eq:proposed}
\end{align}
The first term in Eq. (\ref{eq:proposed}) is KL divergence between the encoder and standard Gaussian distribution,
which can be calculated in a closed form.
The second term is the expectation of the logarithm of the density ratio $q_{\mathbf{\phi}}(\mathbf{z})/p(\mathbf{z})$.
We estimate $q_{\mathbf{\phi}}(\mathbf{z})/p(\mathbf{z})$ with the density ratio trick.
Since the latent variable vector $\mathbf{z}$ is low-dimensional, the density ratio trick works well.

We can estimate the density ratio $q_{\mathbf{\phi}}(\mathbf{z})/p(\mathbf{z})$ as follows.
First, we prepare the samples from $q_{\mathbf{\phi}}(\mathbf{z})$ and samples from $p(\mathbf{z})$.
We can sample from $p(\mathbf{z})$ and $q_{\mathbf{\phi}}(\mathbf{z}\mid\mathbf{x})$ since these distributions are a Gaussian,
and we can also sample from the aggregated posterior $q_{\mathbf{\phi}}(\mathbf{z})$ by using ancestral sampling:
we choose a data point $\mathbf{x}$ from a dataset randomly and sample $\mathbf{z}$ from the encoder given this data point $\mathbf{x}$.
Second, we label $y=1$ to samples from $q_{\mathbf{\phi}}(\mathbf{z})$ and $y=0$ to samples from $p(\mathbf{z})$.
Then, we define $p^{\ast}(\mathbf{z}\mid y)$ as follows:
\begin{align}
  p^{\ast}(\mathbf{z}\mid y)&\equiv
  \begin{cases}
    q_{\mathbf{\phi}}(\mathbf{z}) & (y=1)\\
    p(\mathbf{z}) & (y=0)
  \end{cases}.
\end{align}
Third, we introduce a probabilistic binary classifier $D(\mathbf{z})$ that discriminates between the samples from $q_{\mathbf{\phi}}(\mathbf{z})$ and samples from $p(\mathbf{z})$.
If $D(\mathbf{z})$ can discriminate these samples perfectly,
we can rewrite the density ratio $q_{\mathbf{\phi}}(\mathbf{z})/p(\mathbf{z})$ by using Bayes theorem and $D(\mathbf{z})$ as follows:
\begin{align}
  \frac{q_{\mathbf{\phi}}(\mathbf{z})}{p(\mathbf{z})}&=\frac{p^{\ast}(\mathbf{z}\mid y=1)}{p^{\ast}(\mathbf{z}\mid y=0)}
  =\frac{p^{\ast}(y=0)p^{\ast}(y=1\mid\mathbf{z})}{p^{\ast}(y=1)p^{\ast}(y=0\mid\mathbf{z})}\nonumber\\
  &=\frac{p^{\ast}(y=1\mid\mathbf{z})}{p^{\ast}(y=0\mid\mathbf{z})}
  \equiv\frac{D(\mathbf{z})}{1-D(\mathbf{z})},
\end{align}
where
$p^{\ast}(y=0)$ equals $p^{\ast}(y=1)$ since the number of samples is the same.
We model $D(\mathbf{z})$ by $\sigma(T_{\mathbf{\psi}}(\mathbf{z}))$,
where $T_{\mathbf{\psi}}(\mathbf{z})$ is a neural network with parameter $\psi$ and input $\mathbf{z}$, and $\sigma(\cdot)$ is a sigmoid function.
We train $T_{\mathbf{\psi}}(\mathbf{z})$ to maximize the following objective function:
\begin{multline}
  T^{\ast}(\mathbf{z})=\max_{\mathbf{\psi}}\mathbb{E}_{q_{\mathbf{\phi}}(\mathbf{z})}\left[\ln(\sigma(T_{\mathbf{\psi}}(\mathbf{z})))\right]\\
  +\mathbb{E}_{p(\mathbf{z})}\left[\ln(1-\sigma(T_{\mathbf{\psi}}(\mathbf{z})))\right].
  \label{eq:logistic}
\end{multline}
By using $T^{\ast}(\mathbf{z})$, we can estimate the density ratio $q_{\mathbf{\phi}}(\mathbf{z})/p(\mathbf{z})$ as follows:
\begin{align}
  \frac{q_{\mathbf{\phi}}(\mathbf{z})}{p(\mathbf{z})}=\frac{\sigma(T^{\ast}(\mathbf{z}))}{1-\sigma(T^{\ast}(\mathbf{z}))}\Leftrightarrow T^{\ast}(\mathbf{z})=\ln\frac{q_{\mathbf{\phi}}(\mathbf{z})}{p(\mathbf{z})}.
\end{align}
Therefore, we can estimate the KL divergence with the aggregated posterior $D_{KL}(q_{\mathbf{\phi}}(\mathbf{z}\mid\mathbf{x})\Vert q_{\mathbf{\phi}}(\mathbf{z}))$ by
\begin{multline}
  D_{KL}(q_{\mathbf{\phi}}(\mathbf{z}\mid\mathbf{x})\Vert q_{\mathbf{\phi}}(\mathbf{z}))\\
  =D_{KL}(q_{\mathbf{\phi}}(\mathbf{z}\mid\mathbf{x})\Vert p(\mathbf{z}))-\mathbb{E}_{q_{\mathbf{\phi}}(\mathbf{z}\mid\mathbf{x})}\left[T^{\ast}(\mathbf{z})\right].
  \label{eq:KLD}
\end{multline}

\subsection{Optimization Procedure}
\label{sec:optimization}

From the above discussion,
we obtain the training objective function of the VAE with our implicit optimal prior:
\begin{multline}
  \max_{\mathbf{\theta},\mathbf{\phi}}\int p_{\mathcal{D}}(\mathbf{x})\left\{
  -D_{KL}(q_{\mathbf{\phi}}(\mathbf{z}\mid\mathbf{x})\Vert p(\mathbf{z}))
  \right.\\
  \left.
  +\mathbb{E}_{q_{\mathbf{\phi}}(\mathbf{z}\mid\mathbf{x})}\left[\ln p_{\mathbf{\theta}}(\mathbf{x}\mid\mathbf{z})+T_{\mathbf{\psi}}(\mathbf{z})\right]
  \right\} \mathrm{d}\mathbf{x},
\end{multline}
where $T_{\mathbf{\psi}}(\mathbf{z})$ maximizes the Eq. (\ref{eq:logistic}).
Given a dataset $\mathbf{X}=\left\{\mathbf{x}^{(1)},\ldots,\mathbf{x}^{(N)}\right\}$,
we optimize the Monte Carlo approximation of this objective:
\begin{multline}
  \max_{\mathbf{\theta},\mathbf{\phi}}\frac{1}{N}\sum_{i=1}^{N}\left\{
  -D_{KL}(q_{\mathbf{\phi}}(\mathbf{z}\mid\mathbf{x}^{(i)})\Vert p(\mathbf{z}))
  \right.\\
  \left.
  +\mathbb{E}_{q_{\mathbf{\phi}}(\mathbf{z}\mid\mathbf{x}^{(i)})}\left[\ln p_{\mathbf{\theta}}(\mathbf{x}^{(i)}\mid\mathbf{z})+T_{\mathbf{\psi}}(\mathbf{z})\right]
  \right\},
\end{multline}
and we approximate the expectation term by the reparameterization trick \cite{kingma2013auto}:
\begin{multline}
  \mathbb{E}_{q_{\mathbf{\phi}}(\mathbf{z}\mid\mathbf{x}^{(i)})}\left[\ln p_{\mathbf{\theta}}(\mathbf{x}^{(i)}\mid\mathbf{z})+T_{\mathbf{\psi}}(\mathbf{z})\right]\\
  \simeq\frac{1}{L}\sum_{\ell=1}^{L}\left\{ \ln p_{\mathbf{\theta}}(\mathbf{x}^{(i)}\mid\mathbf{z}^{(i,\ell)})+T_{\mathbf{\psi}}(\mathbf{z}^{(i,\ell)})\right\},
\end{multline}
where $\mathbf{z}^{(i,\ell)}=\mu_{\phi}(\mathbf{x}^{(i)})+\mathbf{\varepsilon}^{(i,\ell)}\odot\sigma_{\phi}(\mathbf{x}^{(i)})$,
$\mathbf{\varepsilon}^{(i,\ell)}$ is a sample drawn from $\mathcal{N}(\mathbf{z}|\mathbf{0},\mathbf{I})$,
$\odot$ is the element-wise product,
and $L$ is the sample size of the reparameterization trick.
Then, the resulting objective function is
\begin{multline}
  \max_{\mathbf{\theta},\mathbf{\phi}}\frac{1}{N}\sum_{i=1}^{N}\Bigl[
  -D_{KL}(q_{\mathbf{\phi}}(\mathbf{z}\mid\mathbf{x}^{(i)})\Vert p(\mathbf{z}))
  \Bigr.\\
  \Bigl.
  +\frac{1}{L}\sum_{\ell=1}^{L}\left\{ \ln p_{\mathbf{\theta}}(\mathbf{x}^{(i)}\mid\mathbf{z}^{(i,\ell)})+T_{\mathbf{\psi}}(\mathbf{z}^{(i,\ell)})\right\}
  \Bigr].
  \label{eq:final_objective}
\end{multline}

We optimize this model with stochastic gradient descent (SGD)~\cite{duchi2011adaptive,zeiler2012adadelta,tieleman2012lecture,kingma2014adam} by iterating a two-step procedure:
we first update $\mathbf{\theta}$ and $\mathbf{\phi}$ to maximize Eq. (\ref{eq:final_objective}) with fixed $\mathbf{\psi}$
and next update $\mathbf{\psi}$ to maximize the Monte Carlo approximation of Eq. (\ref{eq:logistic}) with fixed $\mathbf{\theta}$ and $\mathbf{\phi}$, as follows:
\begin{multline}
  \max_{\mathbf{\psi}}\frac{1}{M}\sum_{i=1}^{M}\ln(\sigma(T_{\mathbf{\psi}}(\mathbf{z}_{1}^{(i)})))\\
  +\frac{1}{M}\sum_{j=1}^{M}\ln(1-\sigma(T_{\mathbf{\psi}}(\mathbf{z}_{0}^{(j)}))),
  \label{eq:mc_logistic}
\end{multline}
where $\mathbf{z}_{1}^{(i)}$ is a sample drawn from $q_{\mathbf{\phi}}(\mathbf{z})$,
$\mathbf{z}_{0}^{(j)}$ is a sample drawn from $p(\mathbf{z})$,
and $M$ is the sampling size of Monte Carlo approximation.
Note that we need to compute the gradient of $T_{\mathbf{\psi}}(\mathbf{z})$ with respect to $\mathbf{\phi}$ in the optimization of Eq. (\ref{eq:final_objective}) since $T_{\mathbf{\psi}}(\mathbf{z})$ models $\ln q_{\mathbf{\phi}}(\mathbf{z})/p(\mathbf{z})$.
However, when $T_{\mathbf{\psi}}(\mathbf{z})$ equals $T^{\ast}(\mathbf{z})$,
the expectation of this gradient becomes zero, as follows:
\begin{align}
  &\mathbb{E}_{p_{\mathcal{D}}(\mathbf{x})q_{\mathbf{\phi}}(\mathbf{z}\mid\mathbf{x})}\left[\nabla_{\mathbf{\phi}}T^{\ast}(\mathbf{z})\right]
  =\mathbb{E}_{q_{\mathbf{\phi}}(\mathbf{z})}\left[\nabla_{\mathbf{\phi}}\ln q_{\mathbf{\phi}}(\mathbf{z})\right]\nonumber\\
  &=\int q_{\mathbf{\phi}}(\mathbf{z})\frac{\nabla_{\mathbf{\phi}}q_{\mathbf{\phi}}(\mathbf{z})}{q_{\mathbf{\phi}}(\mathbf{z})}\mathrm{d}\mathbf{z}
  =\nabla_{\mathbf{\phi}}\int q_{\mathbf{\phi}}(\mathbf{z})\mathrm{d}\mathbf{z}
  =\nabla_{\mathbf{\phi}}1
  =0.
\end{align}
Therefore, we ignore this gradient in the optimization
\footnote{There is almost the same discussion in \cite{mescheder2017adversarial}.}.
We also note that $T_{\mathbf{\psi}}(\mathbf{z})$ is likely to overfit to the log density ratio between the empirical aggregated posterior (Eq. (\ref{eq:empirical})) and the standard Gaussian distribution.
As mentioned in Section \ref{sec:vamp},
this over-fitting also incurs over-fitting of the VAE \cite{tomczak2017vae}.
Therefore, we use the regularization techniques such as dropout \cite{srivastava2014dropout} for $T_{\mathbf{\psi}}(\mathbf{z})$,
which prevents it from over-fitting.
We train $\mathbf{\psi}$ more than $\mathbf{\theta}$ and $\mathbf{\phi}$:
if we update $\mathbf{\theta}$ and $\mathbf{\phi}$ for $J_{1}$ steps,
we update $\mathbf{\psi}$ for $J_{2}$ steps, where $J_{2}$ is larger than $J_{1}$.
Algorithm \ref{alg:proposed} shows the pseudo code of the optimization procedure of this model, where $K$ is the minibatch size of SGD.

\begin{algorithm}[tb]
  \caption{VAE with Implicit Optimal Priors}
  \begin{algorithmic}[1]
    \While{not converged}
      \For{$J_{1}$ steps}
        \State{Sample minibatch $\left\{\mathbf{x}^{(1)},\ldots,\mathbf{x}^{(K)}\right\}$ from $\mathbf{X}$}
        \State{Compute the gradients of Eq. (\ref{eq:final_objective}) w.r.t. $\mathbf{\theta}$ and $\mathbf{\phi}$}
        \State{Update $\mathbf{\theta}$ and $\mathbf{\phi}$ with their gradients}
      \EndFor
      \For{$J_{2}$ steps}
        \State{Sample minibatch $\left\{ \mathbf{z}_{0}^{(1)},\ldots,\mathbf{z}_{0}^{(K)}\right\} $ from $p(\mathbf{z})$}
        \State{Sample minibatch $\left\{ \mathbf{z}_{1}^{(1)},\ldots,\mathbf{z}_{1}^{(K)}\right\} $ from $q_{\mathbf{\phi}}(\mathbf{z})$}
        \State{Compute the gradient of Eq. (\ref{eq:mc_logistic}) w.r.t. $\mathbf{\psi}$}
        \State{Update $\mathbf{\psi}$ with its gradient}
      \EndFor
    \EndWhile
  \end{algorithmic}
  \label{alg:proposed}
\end{algorithm}

\section{Related Work}
\label{sec:related}

For improving the density estimation performance of the VAE,
numerous works have focused on the regularization effect of the KL divergence between the encoder and the prior.
These works improve either the encoder or the prior.

First, we focus on the works about the prior.
Although the optimal prior for the VAE is the aggregated posterior,
the KL divergence with the aggregated posterior cannot be calculated in a closed form.
As described in Section \ref{sec:vamp}, the VampPrior \cite{tomczak2017vae} has been presented to solve this problem.
However, it has sensitive hyperparameters such as the number of mixtures $K$.
Since the VampPrior requires a heavy computational cost, these hyperparameters are difficult to tune.
In contrast to this, our approach can estimate the KL divergence more easily and robustly than the VampPrior since it does not need to model the aggregated posterior explicitly.
In addition, since the computational cost of our approach is much more lightweight than that of VampPrior,
the hyperparameters of our approach are easier to tune than those of VampPrior.

There are approaches on improving the prior other than the aggregated posterior.
For example, non-parametric Bayesian distribution \cite{nalisnick2017stick} and hyperspherical distribution \cite{davidson2018hyperspherical} are used for the prior.
These approaches aim to obtain the useful and interpretable latent representation rather than improving the density estimation performance,
which is opposite to our purpose.
We should mention the disadvantage of our approach compared with these approaches.
Since our prior is implicit, we cannot sample from our prior directly.
Instead, we can sample from the aggregated posterior, which our implicit prior models, by using ancestral sampling.
That is, when we sample from the prior, we need to prepare a data point.

Next, we focus on the works about the encoder.
To improve the density estimation performance, these works increase the flexibility of the encoder.
The normalizing flow \cite{rezende2015variational,kingma2016improved,huang2018neural} is one of the main approaches,
which applies a sequence of invertible transformations to the latent variable vector until a desired level of flexibility is attained.
Our approach is orthogonal to the normalizing flow and can be used together with it.

The similar approaches to ours are the adversarial variational Bayes (AVB) \cite{mescheder2017adversarial} and the adversarial autoencoders (AAE) \cite{makhzani2015adversarial,tolstikhin2017wasserstein}.
These approaches use the implicit encoder network, which takes as input a data point $\mathbf{x}$ and Gaussian random noise and produces a latent variable vector $\mathbf{z}$.
Since the implicit encoder does not assume the distribution type, it can become a very flexible distribution.
In these approaches, the standard Gaussian distribution is used for the prior.
Although the KL divergence between the implicit encoder and the standard Gaussian prior $D_{KL}(q_{\mathbf{\phi}}(\mathbf{z}\mid\mathbf{x})\Vert p(\mathbf{z}))$ cannot be calculated in a closed form,
the AVB estimates this KL divergence by using the density ratio trick.
However, this estimation does not work well with high-dimensional datasets
since this KL divergence also becomes a high-dimensional density ratio \cite{rosca2018distribution}.
Our approach can avoid this problem since we use the density ratio trick in a low dimension.
The AAE is an expansion of the Autoencoder rather than the VAE.
The AAE regularizes the aggregated posterior to be close to the standard Gaussian prior by minimizing the KL divergence $D_{KL}(q_{\mathbf{\phi}}(\mathbf{z})\Vert p(\mathbf{z}))$.
The AAE also uses the density ratio trick to estimate this KL divergence,
and this works well since this KL divergence is a low-dimensional density ratio.
However, the AAE cannot estimate the probability of a data point.
Our approach is based on the VAE, and can estimate the probability of a data point.

\section{Experiments}
\label{sec:experiment}

In this section, we experimentally evaluate the density estimation performance of our approach.

\subsection{Data}

We used five datasets:
OneHot \cite{mescheder2017adversarial},
MNIST \cite{salakhutdinov2008quantitative},
OMNIGLOT \cite{burda2015importance},
FreyFaces\footnote{This dataset is available at \url{https://cs.nyu.edu/~roweis/data/frey_rawface.mat}},
and Histopathology \cite{tomczak2016improving}.
OneHot consists of only four-dimensional one hot vectors:
$(1,0,0,0)^{\mathrm{T}}$,
$(0,1,0,0)^{\mathrm{T}}$,
$(0,0,1,0)^{\mathrm{T}}$,
and $(0,0,0,1)^{\mathrm{T}}$.
This simple dataset is useful for observing the posterior of the latent variable,
which is used in \cite{mescheder2017adversarial}.
MNIST and OMNIGLOT are binary image datasets, and FreyFaces and Histopathology are grayscale image datasets.
These image datasets are useful for measuring the density estimation performance,
which are used in \cite{tomczak2017vae}.
The number and the dimensions of data points of the five datasets are listed in Table \ref{tab:dataset_detail}.

\begin{table}[t]
  \begin{center}
    {\tabcolsep=0.2em
    \caption{Number and dimensions of datasets}
    \begin{tabular}{lrrrr}
      \toprule
      {}             & Dimension & Train size & Valid size & Test size \\
      \midrule
      OneHot         & 4         & 1,000      & 100        & 1,000     \\
      MNIST          & 784       & 50,000     & 10,000     & 10,000    \\
      OMNIGLOT       & 784       & 23,000     & 1,345      & 8,070     \\
      FreyFaces      & 560       & 1,565      & 200        & 200       \\
      Histopathology & 784       & 6,800      & 2,000      & 2,000     \\
      \bottomrule
      \end{tabular}
      \label{tab:dataset_detail}}
  \end{center}
\end{table}

\begin{figure*}[tb]
  \begin{center}
    \begin{minipage}{0.245\hsize}
      \includegraphics[width=\hsize]{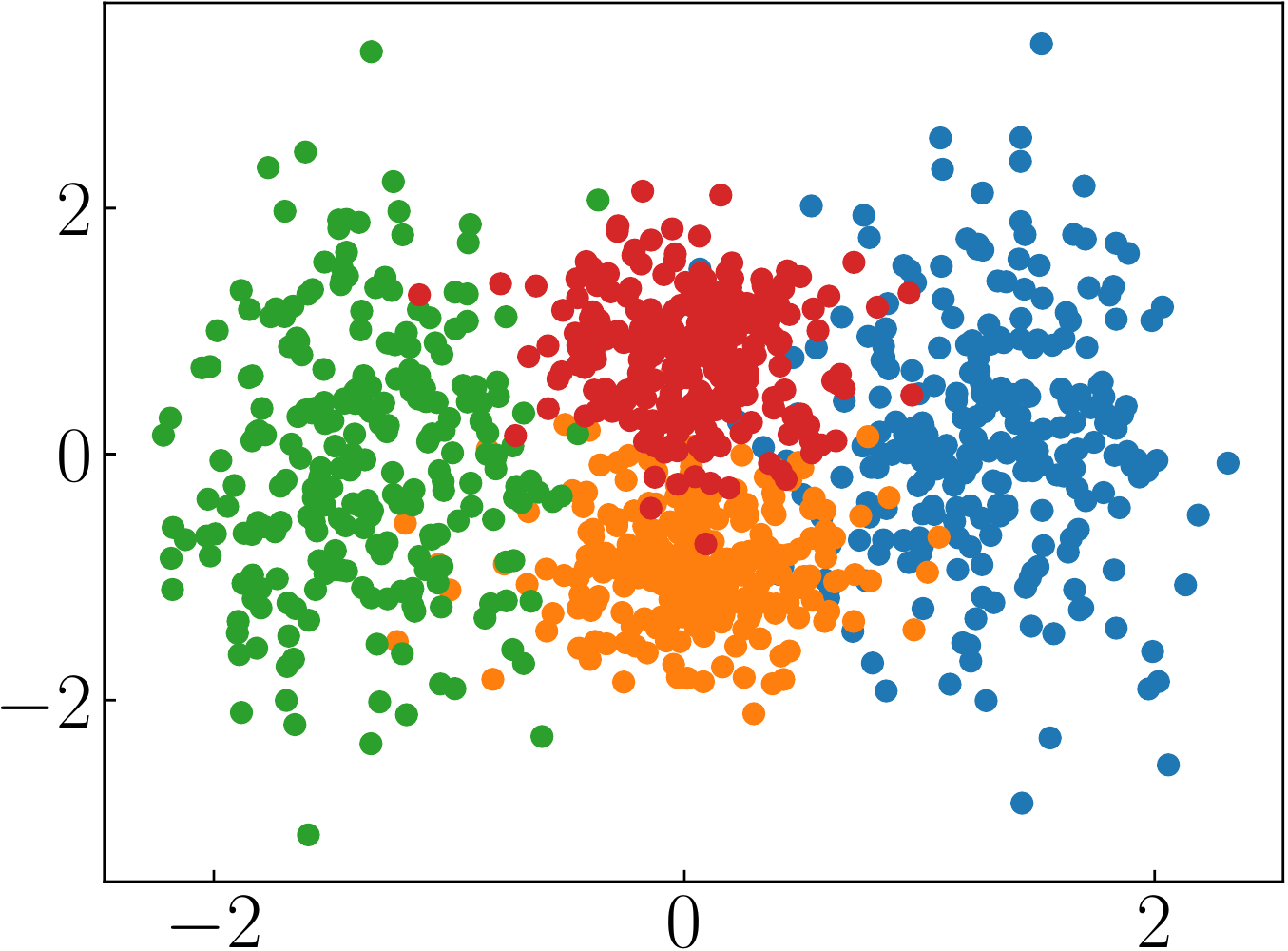}
      \subcaption{Standard VAE.}
      \label{fig:onehot_normal_latent}
    \end{minipage}
    \begin{minipage}{0.245\hsize}
      \includegraphics[width=\hsize]{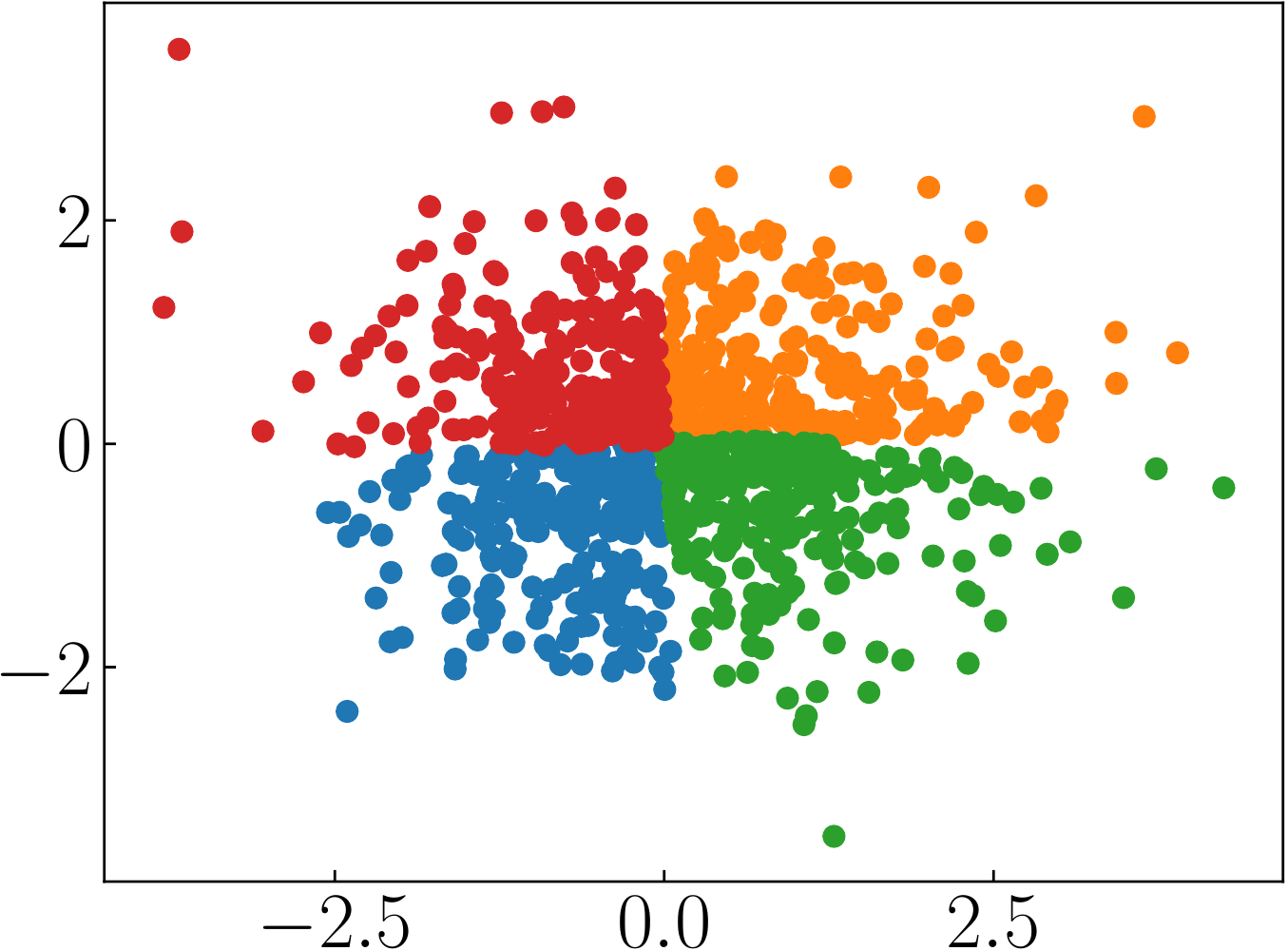}
      \subcaption{AVB.}
      \label{fig:onehot_avb_latent}
    \end{minipage}
    \begin{minipage}{0.245\hsize}
      \includegraphics[width=\hsize]{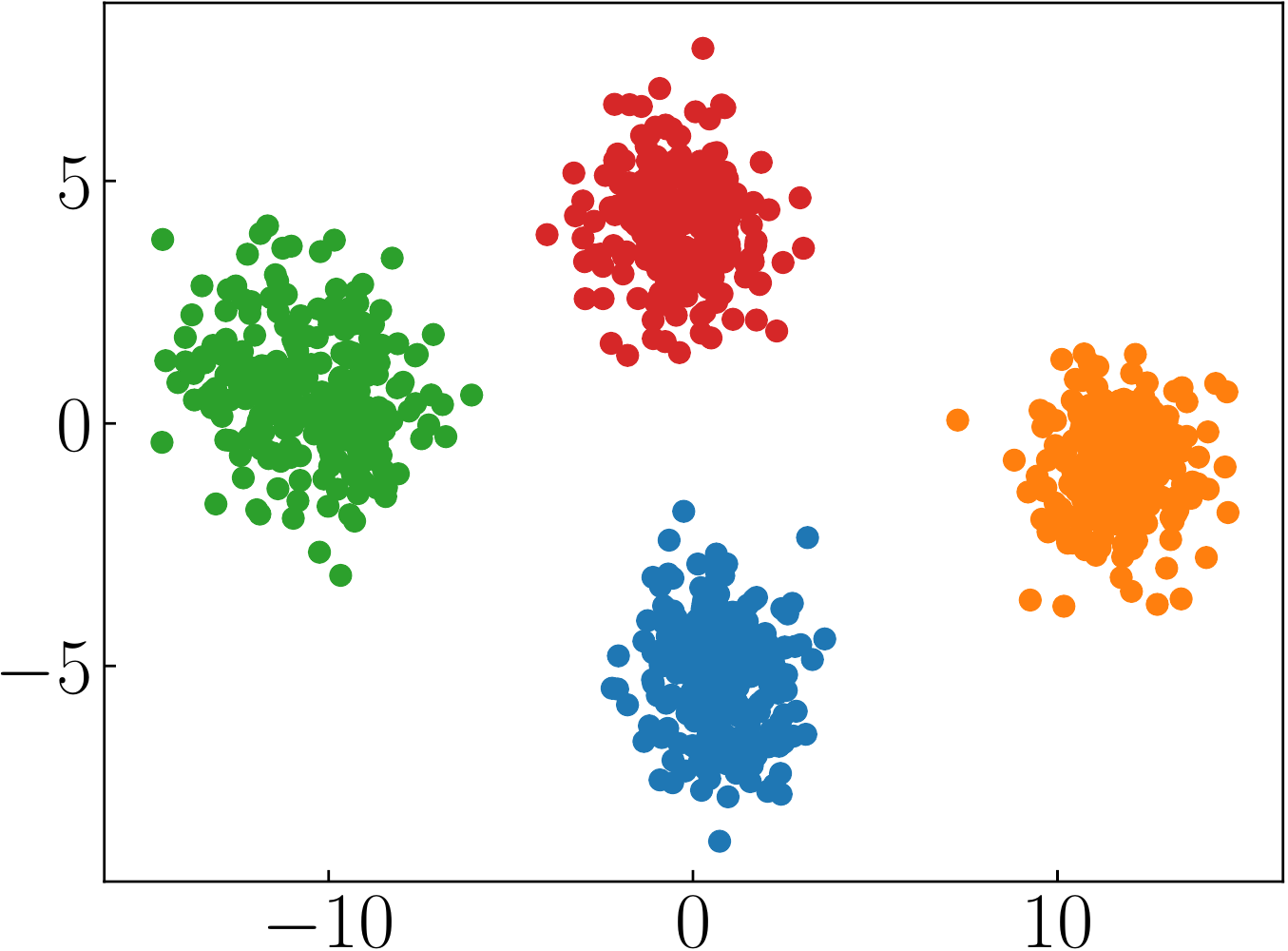}
      \subcaption{VAE with VampPrior.}
      \label{fig:onehot_vamp_latent}
    \end{minipage}
    \begin{minipage}{0.245\hsize}
      \includegraphics[width=\hsize]{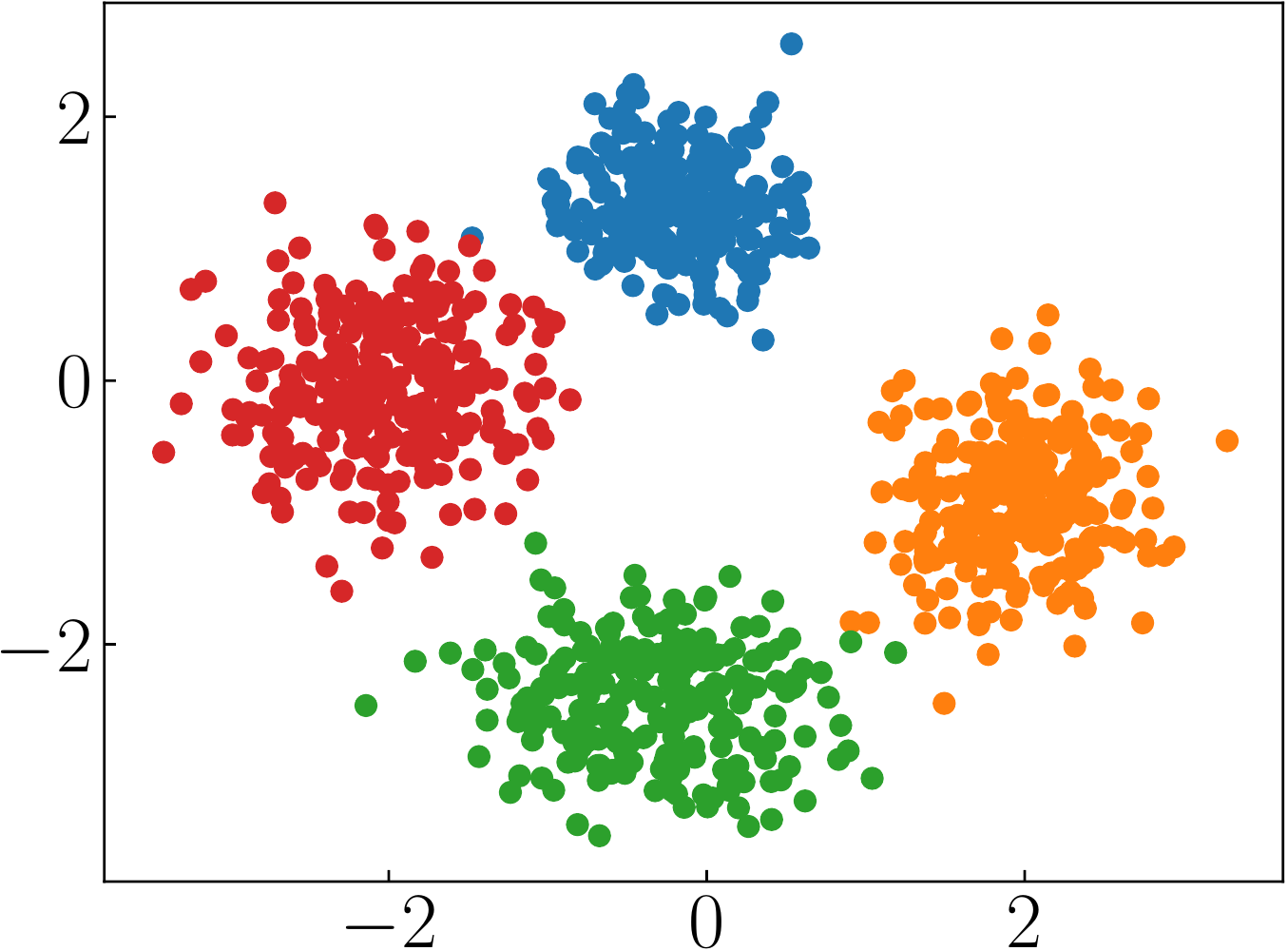}
      \subcaption{Proposed method.}
      \label{fig:onehot_iop_latent}
    \end{minipage}
  \end{center}
  \caption{
    Comparison of posteriors of latent variable on OneHot.
    We plotted samples drawn from $q_{\mathbf{\phi}}(\mathbf{z}\mid\mathbf{x})$,
    where $\mathbf{x}$ is a one hot vector: $(1,0,0,0)^{\mathrm{T}}$, $(0,1,0,0)^{\mathrm{T}}$, $(0,0,1,0)^{\mathrm{T}}$, or $(0,0,0,1)^{\mathrm{T}}$.
    We used test data for this sampling.
    Samples in each color correspond to each latent representation of one hot vectors.
    (a) Standard VAE (VAE with standard Gaussian prior).
    (b) AVB.
    (c) VAE with VampPrior.
    (d) Proposed method.
  }
\end{figure*}

\begin{figure*}[tb]
  \begin{center}
    \begin{minipage}{0.245\hsize}
      \includegraphics[width=\hsize]{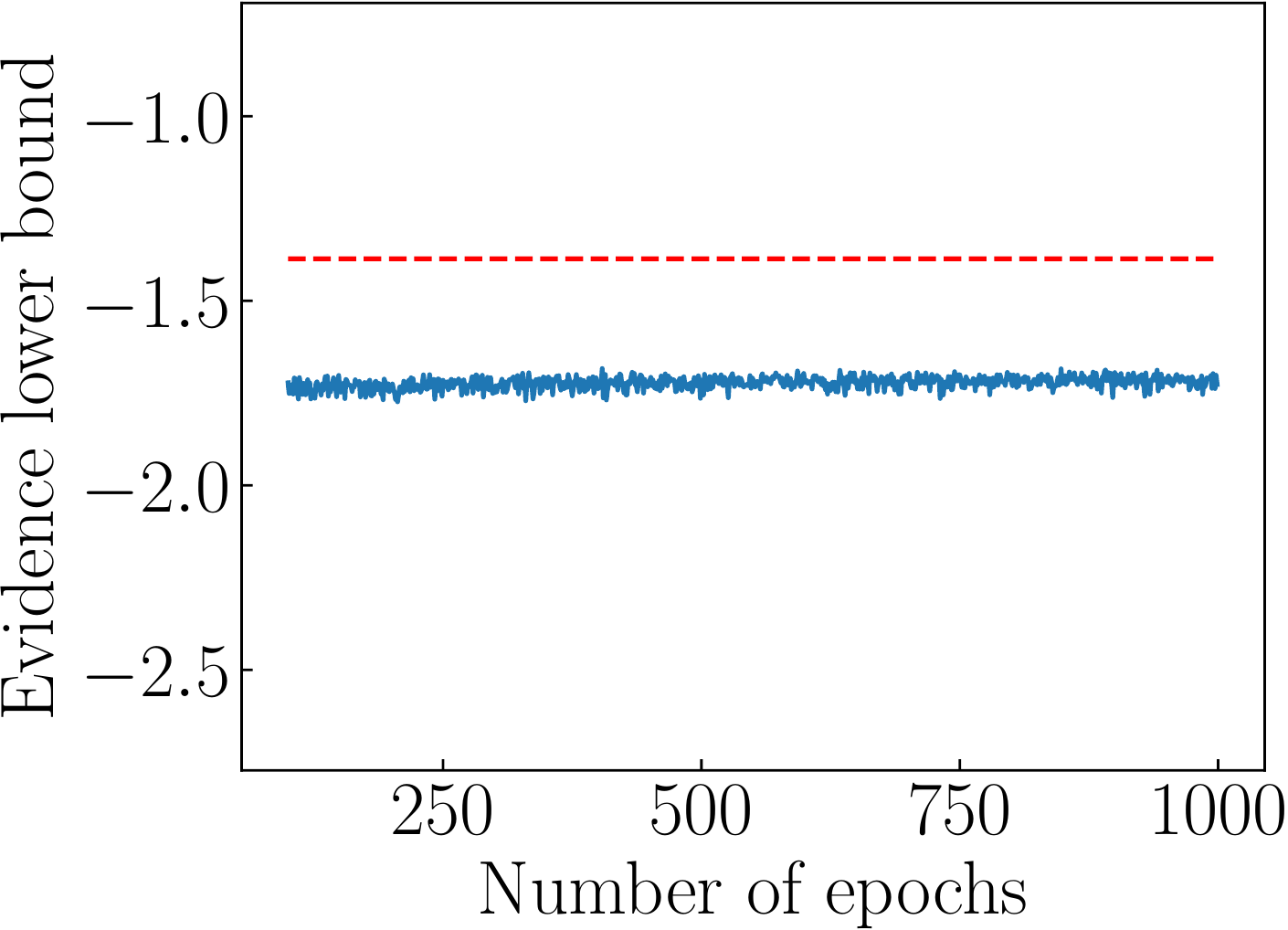}
      \subcaption{Standard VAE.}
      \label{fig:onehot_normal_elbo}
    \end{minipage}
    \begin{minipage}{0.245\hsize}
      \includegraphics[width=\hsize]{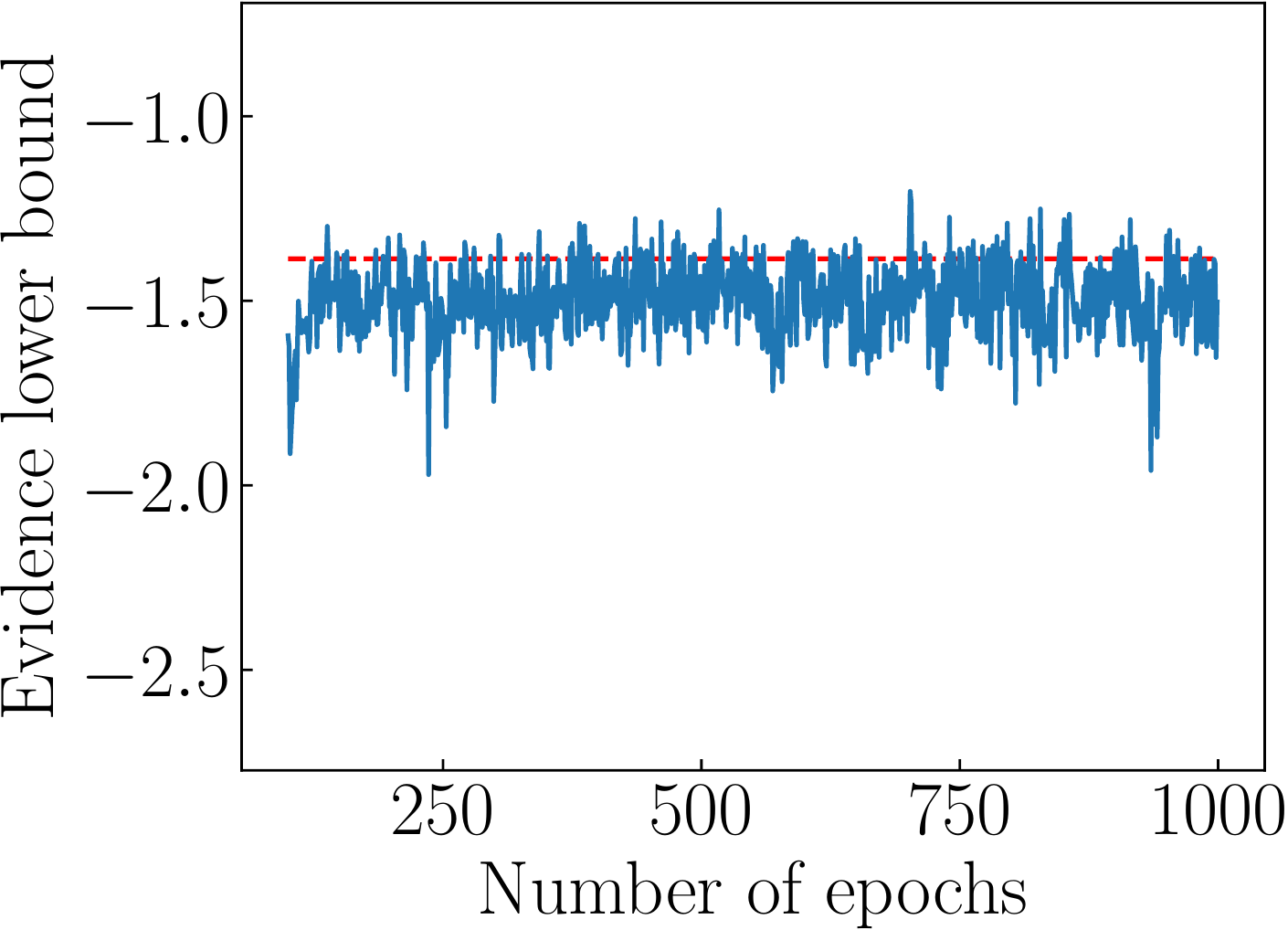}
      \subcaption{AVB.}
      \label{fig:onehot_avb_elbo}
    \end{minipage}
    \begin{minipage}{0.245\hsize}
      \includegraphics[width=\hsize]{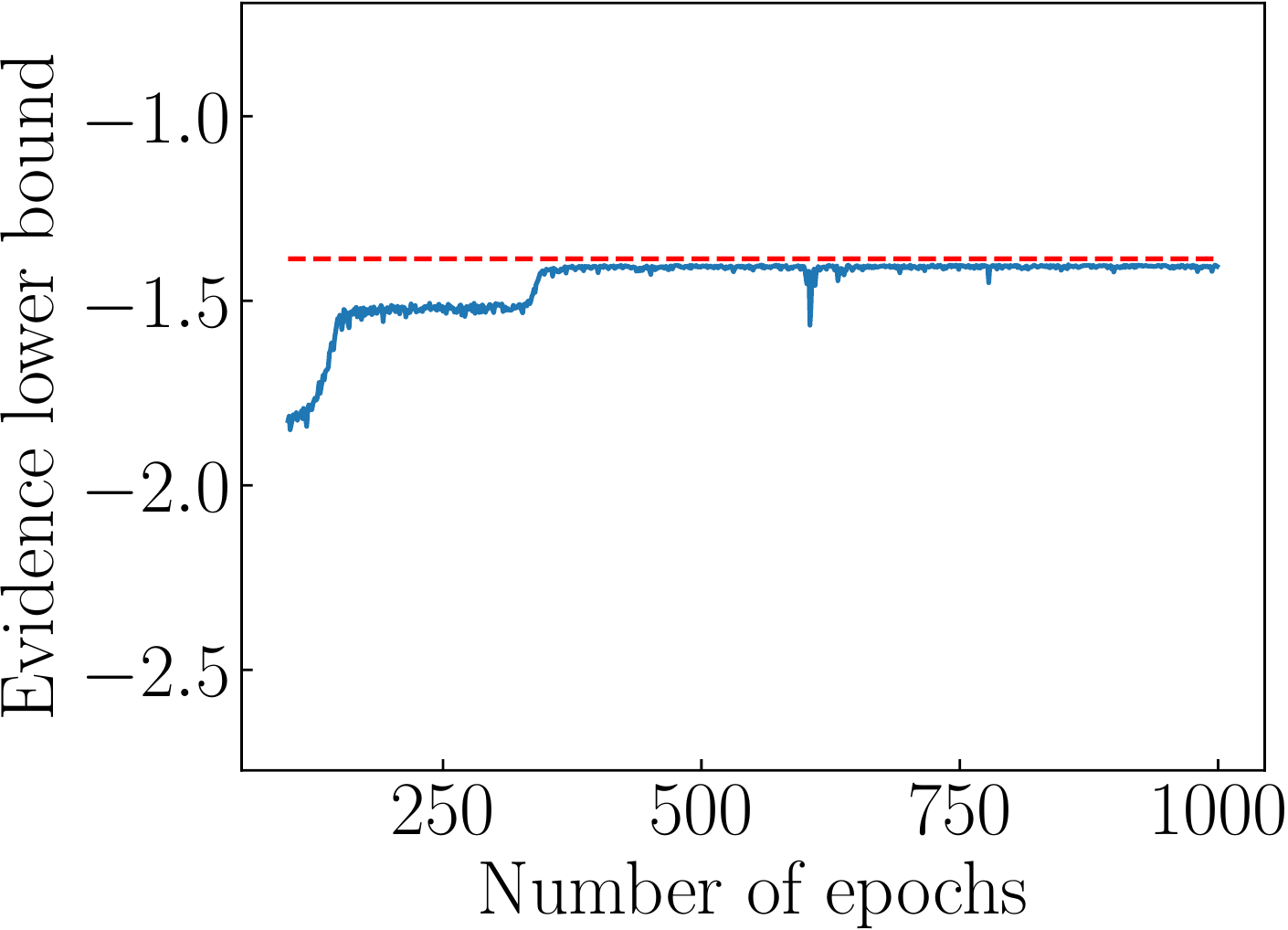}
      \subcaption{VAE with VampPrior.}
      \label{fig:onehot_vamp_elbo}
    \end{minipage}
    \begin{minipage}{0.245\hsize}
      \includegraphics[width=\hsize]{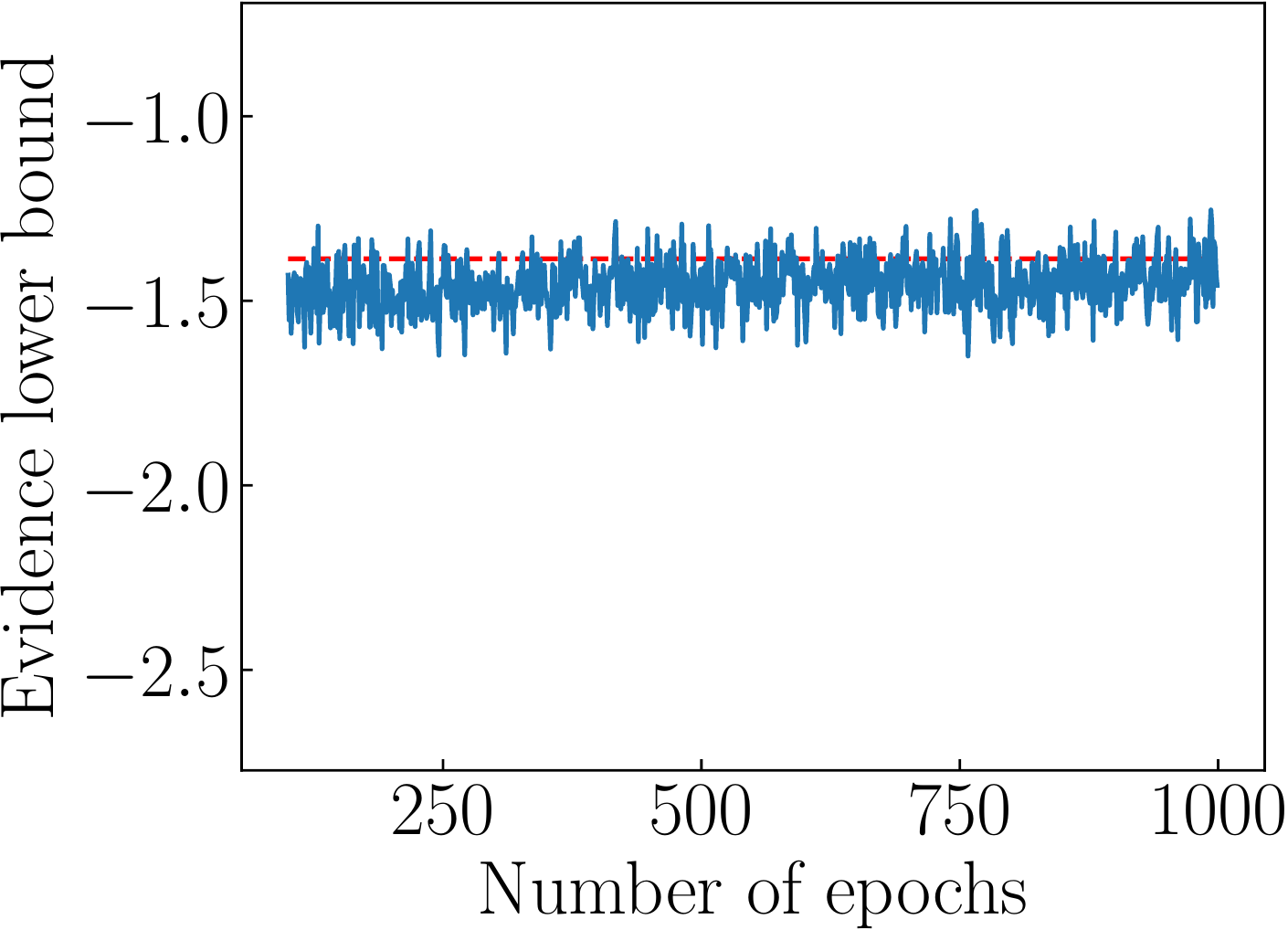}
      \subcaption{Proposed method.}
      \label{fig:onehot_iop_elbo}
    \end{minipage}
  \end{center}
  \caption{
    Comparison of the evidence lower bound (ELBO) with validation data on OneHot.
    We plotted the ELBO from 100 to 1,000 epochs since we used warm-up for the first 100 epochs.
    The optimal log-likelihood on this dataset is $-\ln(4) \approx -−1.386$.
    We plotted this value by a dashed line for comparison.
    (a) Standard VAE (VAE with standard Gaussian prior).
    (b) AVB.
    (c) VAE with VampPrior.
    (d) Proposed method.
  }
\end{figure*}

\begin{table*}[t]
  \begin{center}
    \caption{
      Comparison of test log-likelihoods on four image datasets.
    }
    \begin{tabular}{lrrrr}
      \toprule
                         & MNIST                                  & OMNIGLOT                                & FreyFaces                               & Histopathology                          \\
      \midrule
      Standard VAE       & -85.84 $\pm$ 0.07                      & -111.39 $\pm$ 0.11                      & 1382.53 $\pm$ 3.57                      & 1081.53 $\pm$ 0.70                      \\
      VAE with VampPrior & -83.90 $\pm$ 0.08                      & -110.53 $\pm$ 0.09                      & \bfseries{1392.62 $\pm$ 6.25}           & 1083.11 $\pm$ 2.10                      \\
      Proposed method    & $\approx$ \bfseries{-83.21 $\pm$ 0.13} & $\approx$ \bfseries{-108.48 $\pm$ 0.16} & $\approx$ \bfseries{1396.27 $\pm$ 2.75} & $\approx$ \bfseries{1087.42 $\pm$ 0.60} \\
      \bottomrule
    \end{tabular}
    \label{tab:test_scores}
  \end{center}
\end{table*}

\begin{figure}[tb]
  \begin{center}
    \includegraphics[width=0.8\columnwidth]{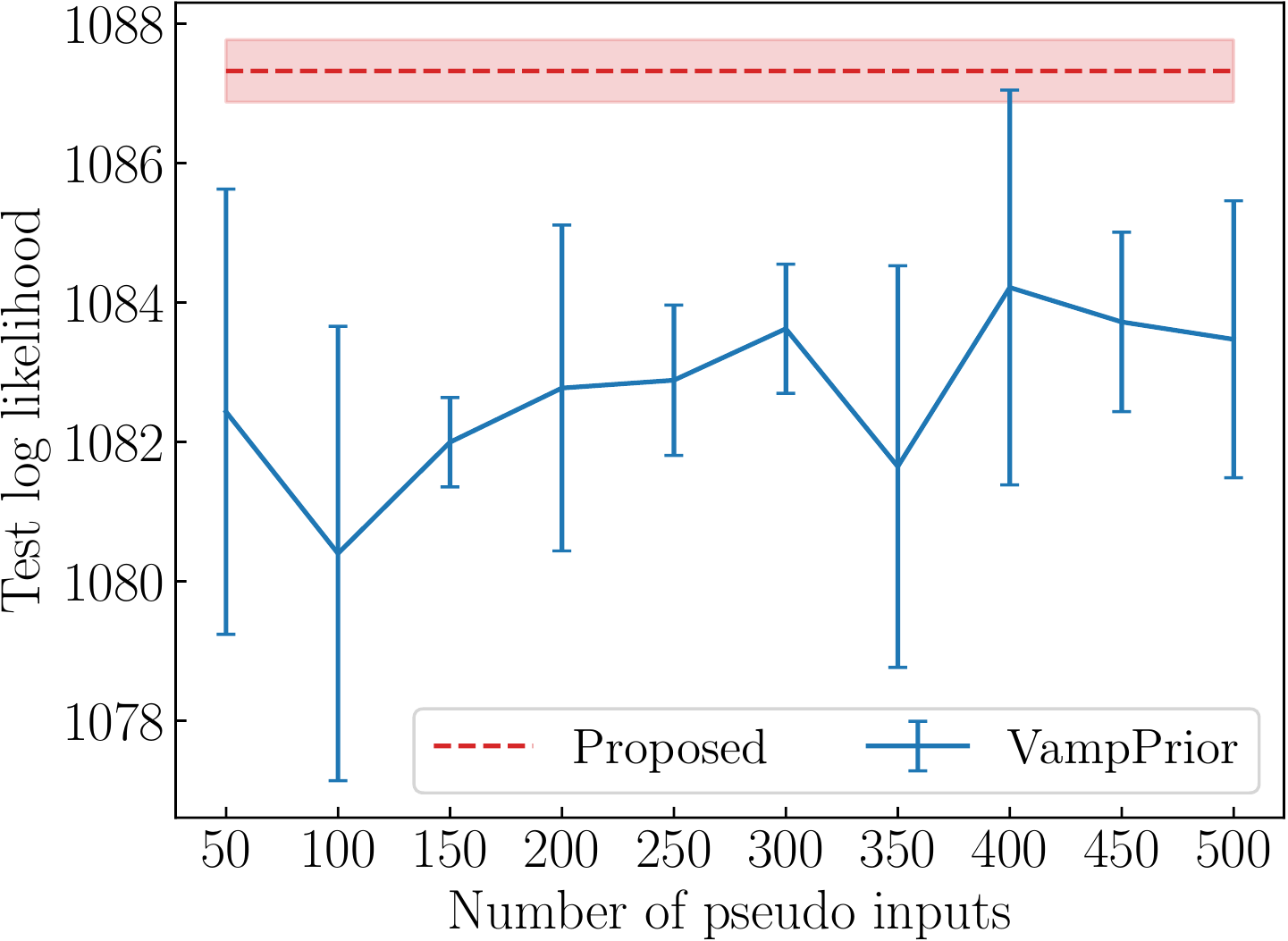}
    \caption{
      Relationship between the test log-likelihoods and number of pseudo inputs of VampPrior on Histopathology.
      We plotted the test log-likelihoods of our approach by a dashed line for comparison.
      The semi-transparent area and error bar represent standard deviations.
    }
    \label{fig:vamp_tuning}
  \end{center}
\end{figure}

\subsection{Setup}
\label{sec:setup}

We compared our implicit optimal prior with standard Gaussian prior and VampPrior.
We set the dimensions of the latent variable vector to 2 for OneHot,
and 40 for other datasets.
We used two-layer neural networks (500 hidden units per layer) for the encoder, the decoder, and the density ratio estimator.
We used the gating mechanism \cite{dauphin2016language} for the encoder and the decoder
and used a hyperbolic tangent as the activation function for the density ratio estimator.
We initialized the weights of these neural networks in accordance with the method in \cite{glorot2010understanding}.
We used a Gaussian distribution as the encoder.
As the decoder, we used a Bernoulli distribution for OneHot, MNIST, and OMNIGLOT
and used a Gaussian distribution for FreyFaces and Histopathology,
means of which were constrained to the interval $\left[0,1\right]$ by using a sigmoid function.
We trained all methods by using Adam \cite{kingma2014adam} with a mini-batch size of 100 and learning rate in $\left[10^{-4}, 10^{-3}\right]$.
We set the maximum number of epochs to 1,000 and used early-stopping \cite{Goodfellow-et-al-2016} on the basis of validation data.
We set the sample size of the reparameterization trick to $L=1$.
In addition, we used warm-up \cite{bowman2015generating} for the first 100 epochs of Adam.
For MNIST and OMNIGLOT, we used dynamic binarization \cite{salakhutdinov2008quantitative} during the training of VAE to avoid over-fitting.
For image datasets,
we calculated the log marginal likelihood of the test data by using the importance sampling \cite{burda2015importance}.
We set the sample size of the importance sampling to 10.
We ran all experiments eight times each.

With VampPrior,
we set the number of mixtures $K$ to 50 for OneHot,
500 for MNIST, FreyFaces, and Histopathology,
and 1,000 for OMNIGLOT.
In addition, for image datasets,
we used a clipped relu function that equals $\min(\max(x,0),1)$ to scale the pseudo inputs in $\left[0,1\right]$ since the range of data points of these datasets is $\left[0,1\right]$
\footnote{We referred to \url{https://github.com/jmtomczak/vae_vampprior}}.

With our approach,
we used dropout \cite{srivastava2014dropout} in the training of the density ratio estimator since it is likely to over-fit.
We set the keep probability of dropout to 50\%.
We updated the parameter of the density ratio estimator: $\mathbf{\psi}$ for 10 epochs
during the updating of the parameters of VAE: $\mathbf{\theta}$ and $\mathbf{\phi}$ for one epoch.
We set the sampling size of Monte Carlo approximation in Eq. (\ref{eq:mc_logistic}) to $M=N$.

In addition, we compared our approach with adversarial variational Bayes (AVB) on OneHot.
We set the dimension of the Gaussian random noise input of AVB to 10,
and other settings are almost the same as those for our approach.

\subsection{Results}
\label{sec:evaluation}

Figures \ref{fig:onehot_normal_latent}--\ref{fig:onehot_iop_latent} show the posteriors of latent variable of each approach on OneHot,
and Figures \ref{fig:onehot_normal_elbo}--\ref{fig:onehot_iop_elbo} show the evidence lower bound of each approach on OneHot.

These results show the difference between these approaches.
We can see that the evidence lower bound (ELBO) of the standard VAE (VAE with standard Gaussian prior) on OneHot was worse than the optimal log-likelihood on this dataset: $-\ln(4) \approx -−1.386$.
The over-regularization incurred by the standard Gaussian prior can be given as a reason.
The posteriors were overlapped, and it became difficult to discriminate between samples from these posteriors.
Hence, the decoder became confused when reconstructing.
This caused the poor density estimation performance.

On the other hand,
the ELBOs of AVB, VAE with VampPrior, and our approach are much closer to the optimal log-likelihood than the standard VAE.
We note that the ELBOs of the AVB and our approach are the estimated values,
and that these approaches may overestimate the ELBO on OneHot since the training data and validation data of OneHot are the same.
First, we focus on the AVB.
Although there is still the strong regularization by the standard Gaussian prior,
the posteriors barely overlapped,
and the data point was easy to reconstruct from the latent representation.
The reason is that the implicit encoder network of AVB can learn complex posterior distributions.
Next, we focus on the VAE with VampPrior and our approach.
The VampPrior and our implicit optimal prior model the aggregated posterior that is the optimal prior for the VAE.
These priors made the posteriors of these approaches different from each other,
and the data point was easy to reconstruct from the latent representation.

Table \ref{tab:test_scores} compares the test log-likelihoods on four image datasets.
We used bold to highlight the best result and the results that are not statistically different from the best result according to a pair-wise $t$-test.
We used 5\% as the p-value.
We did not compare with AVB since the estimated log marginal likelihood of AVB with high-dimensional datasets such as images is not accurate \cite{rosca2018distribution}.

First, we focus on the VampPrior.
We can see that test log-likelihoods of VampPrior are better than those of standard VAE.
However, we found two drawbacks with the VampPrior.
One is that the pseudo inputs of VampPrior are difficult to optimize.
For example, the pseudo inputs have an initial value dependence.
Although the warm-up helps in solving this problem,
it seems difficult to solve completely.
The other is that the number of mixtures $K$ is a sensitive hyperparameter.
Figure \ref{fig:vamp_tuning} shows the test log-likelihoods with various $K$ on Histopathology.
The high standard deviation of the VampPrior indicates its high dependence of the pseudo input initial values.
In addition, even though we choose the optimal K, the test log-likelihood of the VampPrior is worse than that of our approach.

Next, we focus on our approach.
Our approach obtained the equal to or better density estimation performance than the VampPrior.
Since our approach models the aggregated posterior implicitly,
it can estimate the KL divergence more easily and robustly than the VampPrior.
In addition, it has a much more lightweight computational cost than the VampPrior.
In the training phase on MNIST, our approach was almost $2.83$ times faster than the VampPrior.
Therefore, although our approach has as many hyperparameters, like the neural architecture of the density ratio estimator, as the VampPrior,
these hyperparameters are easier to tune than those of the VampPrior.

These results indicate that our implicit optimal prior is a good alternative to the VampPrior:
our implicit optimal prior can be optimized easily and robustly,
and its density estimation performance is equal to or better than that of the VAE with the VampPrior.

\section{Conclusion}
\label{sec:conclusion}

In this paper, we proposed the variational autoencoder (VAE) with implicit optimal priors.
Although the standard Gaussian distribution is usually used for the prior,
this simple prior incurs over-regularization, which is one of the causes of poor density estimation performance.
To improve the density estimation performance,
the aggregated posterior has been introduced as a sophisticated prior,
which is optimal in terms of maximizing the training objective function of VAE.
However, Kullback Leibler (KL) divergence between the encoder and the aggregated posterior cannot be calculated in a closed form,
which prevents us from using this optimal prior.
Even though explicit modeling of the aggregated posterior has been tried,
this optimal prior is difficult to model explicitly.

With the proposed method, we introduced the density ratio trick for estimating this KL divergence directly.
Since the density ratio trick can estimate the density ratio between two distributions without modeling each distribution explicitly,
there is no need to model the aggregated posterior explicitly.
Although the density ratio trick is useful, it does not work well in a high dimension.
Unfortunately, the KL divergence between the encoder and the aggregated posterior is high-dimensional.
Hence, we rewrite the KL divergence into the sum of two terms:
the KL divergence between the encoder and the standard Gaussian distribution that can be calculated in a closed form,
and the low-dimensional density ratio between the aggregated posterior and the standard Gaussian distribution,
to which the density ratio trick is applied.
We experimentally showed the high density estimation performance of the VAE with this implicit optimal prior.

\bibliographystyle{aaai}
\bibliography{aaai19}

\end{document}